\documentclass{svproc}
\usepackage{url}
\usepackage{times}
\usepackage{epsfig}
\usepackage{graphicx}
\usepackage{amsmath}
\usepackage{amssymb}
\usepackage{multirow}
\usepackage{array}

\begin{document}
\mainmatter              
\title{A Comparison and Strategy of Semantic Segmentation on Remote Sensing Images}
\titlerunning{Semantic Segmentation}  
%
\author{Junxing Hu\inst{1 2} \and Ling Li\inst{1 2}\and Yijun Lin\inst{1 2} \and Fengge Wu\inst{2} \and Junsuo Zhao\inst{2}}
\authorrunning{Junxing Hu et al.} 
%
%
\institute{University of Chinese Academy of Sciences, Beijing, China,\\
\email{junxing2017@iscas.ac.cn}
\and
Institute of Software Chinese Academy of Sciences (ISCAS),\\
Beijing, China}

\maketitle              

\begin{abstract}

In recent years, with the development of aerospace technology, we use more and more images captured by satellites to obtain information. But a large number of useless raw images, limited data storage resource and poor transmission capability on satellites hinder our use of valuable images. Therefore, it is necessary to deploy an on-orbit semantic segmentation model to filter out useless images before data transmission. In this paper, we present a detailed comparison on the recent deep learning models. Considering the computing environment of satellites, we compare methods from accuracy, parameters and resource consumption on the same public dataset. And we also analyze the relation between them. Based on experimental results, we further propose a viable on-orbit semantic segmentation strategy. It will be deployed on the TianZhi-2 satellite which supports deep learning methods and will be lunched soon.

\end{abstract}
\section{Introduction}

Compared with natural images, remote sensing images always have more kinds of targets with the lower resolution, and irregular-shaped targets impact each other. These bring difficulties to object detection and recognition in remote sensing images.

As one basic method of image understanding, semantic segmentation conducts pixel-level classification of the image. Different from other methods like image classification and object detection, semantic segmentation can produce not only the category, size and quantity of the target, but also accurate boundary and position. Therefore, it is more suitable for processing remote sensing images.

Some traditional semantic segmentation methods are early proposed, such as active contour model~\cite{kass1988snakes}, watershed algorithm~\cite{meyer1990morphological} and graph cuts~\cite{boykov2001interactive}. They usually require artificially setting thresholds or interaction controls, but they are not accurate enough.

In recent years, deep learning method has greatly promoted the development of semantic segmentation. Some deep learning based methods are applied to remote sensing images and achieve good performance. Kampffmeyer et al.~\cite{Kampffmeyer2016Semantic} propose a deep Convolutional Neural Network (CNN) for small objects segmentation in remote sensing images of urban areas. Volpi et al.~\cite{volpi2017dense} use the Fully Convolutional Networks (FCN)~\cite{long2015fully} with 50 layers deep residual networks~\cite{he2016deep} for image segmentation.
There have been some works focusing on the comparison of deep learning based semantic segmentation methods. Garcia-Garcia et al.~\cite{garcia2017review} introduce and compare structures and evaluation results on several natural image datasets for most of the existing semantic segmentation methods. Ball et al.~\cite{ball2017comprehensive} provide applications and challenges for deep learning theories and related tools in remote sensing images.
At present, a comparative research of semantic segmentation models about accuracy, parameters and resource consumption is still lacking for remote sensing images.

TianZhi satellites are designed to verify the feasibility of  using satellite-based intelligent applications, and finally build the space-based intelligent system. With the successful launch of TianZhi-1, the feasibility has been initially verified. Unlike TianZhi-1, TianZhi-2 is equipped with a powerful computing system which supports deep learning algorithms, and it will be launched soon. Based on this platform, we can deploy deep learning models to perform on-orbit semantic segmentation of remote sensing images.

Motivated by such prior studies, we conduct a comparison of recent popular semantic segmentation methods of remote sensing images. The main contributions of this paper can be summarized as follows:

\begin{itemize}
  \item First, we use a new image cropping method for the publicly available remote sensing image dataset. By using the training set of the same size, we fairly compare five efficient semantic segmentation methods in terms of structures, accuracy, parameters and resource consumption. These deep learning based methods are FCN, U-Net~\cite{ronneberger2015u}, SegNet~\cite{badrinarayanan2017segnet}, Pyramid Scene Parsing Network (PSPNet) \cite{zhao2017pyramid} and DeepLab \cite{chen2018deeplab}.

  \item  Second, we propose an on-orbit semantic segmentation strategy for TianZhi-2 based on experimental results.
\end{itemize}

The remainder of this paper is organized as follows. In Section 2, we introduce the comparison of five methods in detail. We describe experiments and discusses results in Section 3. Section 4 talks about the on-orbit semantic segmentation strategy for TianZhi-2. Finally, the paper is concluded in Section 5.

\section{Methods Comparison}
\begin{figure}[t]
\begin{center}
   \includegraphics[width=\linewidth]{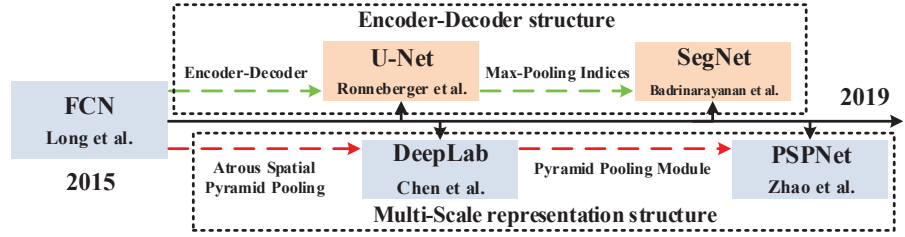}
\end{center}
   \caption{The timeline of five methods. Note that DeepLab was first proposed in 2016 (arXiv), although its official publication was in 2018 (TPAMI).}
\label{timeline}
\end{figure}

Many deep learning based semantic segmentation methods are proposed, and some of them are applied to remote sensing images. We select five most representative methods: FCN, U-Net, SegNet, PSPNet and DeepLab. They and their variants achieve excellent results on some public competitions like the PASCAL VOC-2012 semantic segmentation task~\cite{everingham2015pascal}.
According to the structure, the main deep learning methods for semantic segmentation can be divided into two classes: 1) Encoder-Decoder structure; 2) Multi-Scale representation structure.
In this section, we introduce the five methods based on their relations, respectively. As shown in Figure~\ref{timeline}, we arrange these methods according to the time when they are first proposed.

\subsection{FCN}
FCN~\cite{long2015fully} leads to a rapid increase in the number of semantic segmentation networks. The model transforms all of the fully connected layers to convolutional layers and allows the input image with arbitrary size. In addition, it combines image semantic information from deep layers with shallow layers to produce the final segmentation by using a skip architecture. FCN has been used for remote sensing in~\cite{Kampffmeyer2016Semantic}. The method mainly has three structures as follows:

\textbf{FCN-32s}
It is a basic FCN structure with a single stream, which produces the final segmentation with upsampling at stride 32.

\textbf{FCN-16s}
The method combines predictions from both FCN-32s and the pool4 layer which uses stride 16. It could predict the finer result.

\textbf{FCN-8s}
This model adds predictions from the pool3 layer which is stride 8, and provides further precision. Because remote sensing images contain both coarse, large objects and small, important details, we apply this structure to our experiments in this paper.

\subsection{U-Net}
Built upon FCN, U-Net~\cite{ronneberger2015u} adopts an Encoder-Decoder architecture which consists of a contracting path to capture context and a symmetric expanding path to enable accurate localization. It is originally designed to segment medical images and achieve good results with the fewer training set. Recent years, some studies have shown that U-Net is also suitable for remote sensing images~\cite{huang2018large}, and it has a great potential to be improved. In~\cite{lu2018dual}, Lu et al. propose a dual-resolution U-Net which uses pairs of images as inputs to capture both high and low resolution features.

\subsection{SegNet}
Similar to U-Net, SegNet~\cite{badrinarayanan2017segnet} is also built on an Encoder-Decoder structure. Its encoder network is topologically same as the 13 convolutional layers of the VGG-16~\cite{Simonyan2014VeryDC}. And the decoder network first use max-pooling indices generated from the corresponding encoder to enhance the location information. In~\cite{bischke2017multi}, Bischke et al. use SegNet with a new cascaded multi-task loss to further preserve semantic segmentation boundaries in high resolution satellite images.

\subsection{DeepLab}
DeepLab~\cite{chen2018deeplab} exploits a powerful FCN architecture which mainly uses three components as follows:

\textbf{Atrous Convolution}    The algorithm is originally proposed for computing the undecimated wavelet transform in~\cite{holschneider1990real}. By using the atrous convolution with dilation $d$, the kernel size of a $k \times k$ filter is enlarged to $k_e \times k_e$ as the following formula:
\begin{equation}
k_e= k+(k-1)(d-1)
\end{equation}
The algorithm fills consecutive filter values with $d-1$ zeros to avoid increasing parameters and computation of the model.

\textbf{Atrous Spatial Pyramid Pooling (ASPP)}    ASPP exploits the atrous convolution with different dilation to capture the multiscale information of objects.

\textbf{Fully-Connected Conditional Random Fields (CRFs)}    The fully connected CRFs is integrated into DeepLab to avoid reducing the spatial resolution of feature maps as proposed in~\cite{krahenbuhl2011efficient}. And it greatly improves the localization performance.

\subsection{PSPNet}
Based on DeepLab, PSPNet~\cite{zhao2017pyramid} exploits the pyramid pooling module to aggregate the image global context information with an auxiliary loss. Since DeepLab provides two versions of the model adapted from VGG-16 and ResNet-101, PSPNet can also be applied to VGG and ResNet based network structures, respectively.

Table~\ref{table1} indicates detailed comparisons about methods as above.
%

\begin{table}[h]
\caption{Detailed comparisons of methods. The `-' indicates that the method does not belong to any classes. The `LRP' indicates learning rate policy.}
\begin{center}
\begin{tabular}{l@{\quad}c@{\quad}c@{\quad}c@{\quad}c}
\hline
\rule{0pt}{12pt}Method& Structure& Backbone& LRP& Loss\\[2pt]
\hline
\rule{-3pt}{12pt}
FCN-32s  &  -& VGG-16&  fixed& Cross Entropy\\
FCN-16s  &  Multi-Scale& VGG-16&  fixed& Cross Entropy\\
FCN-8s  &  Multi-Scale&	VGG-16&  fixed&  Cross Entropy\\
U-Net  &  Encoder-Decoder& VGG-16&  step&  Cross Entropy\\
SegNet  &  Encoder-Decoder& VGG-16&  step&  Cross Entropy\\
DeepLab  &  Multi-Scale& VGG-16&  poly&  Cross Entropy\\
DeepLab  &  Multi-Scale& ResNet-101&  poly&  Cross Entropy\\
PSPNet  &  Multi-Scale& VGG-16&  poly&  Cross Entropy\\
PSPNet  &  Multi-Scale& ResNet-101&  poly&  Auxiliary Loss\\[2pt]
\hline
\end{tabular}
\end{center}
\label{table1}
\end{table}


\section{Experiments and Results}
In this section, we first introduce the dataset and the image cropping method. Then, we compare five deep learning models in semantic segmentation with multiple indicators.

\subsection{Datasets}
We evaluate the methods on a public subset of the Inria aerial image labeling benchmark~\cite{maggiori2017can}. The available dataset contains 180 images of size $5000 \times 5000$ at 0.3 m resolution. It is collected from five cities: Austin, Chicago, Kitsap, Vienna and West Tyrol, and each of them has 36 images. As described in~\cite{maggiori2017can}, the first 5 images of each city are used for testing and the other 155 images for training. These pixel-level labeled images contains two classes: building and non-building.

For all images, we extract $512 \times 512$ patches with 12 pixels of overlap between adjacent patches. The size of patch is more suitable for existing semantic segmentation methods and hardware. And the overlap of patches effectively prevent most small objects from being destroyed by image cropping. After extracting, we have 15500 images for training and 2500 images for testing. We use the same training set and test set for all models.

\subsection{Experimental Results}
We compare segmentation performance, parameters, resource consumption and edge prediction of five methods: FCN-8s, U-Net, SegNet, PSPNet and DeepLab v2. All of them are based on VGG structure.
The base learning rate is 0.0001, and it is updated by the ¡°poly¡± policy. All models are trained for 200 epochs and the best results are reported during the training process.


\begin{table}[b]
\caption{Comparisons of evaluation results of motheds on Inria aerial image labeling dataset. `Acc' indicates the accuracy.}
\begin{center}
\begin{tabular}{lp{23pt}<{\centering}p{23pt}<{\centering}p{23pt}<{\centering}p{23pt}<{\centering}p{23pt}<{\centering}p{23pt}<{\centering}p{23pt}<{\centering}p{23pt}<{\centering}p{23pt}<{\centering}p{23pt}<{\centering}p{23pt}<{\centering}p{23pt}<{\centering}}
\hline
\multirow{2}*{Method} & \multicolumn{2}{c}{Austin} & \multicolumn{2}{c}{Chicago} & \multicolumn{2}{c}{Kitsap} & \multicolumn{2}{c}{West Tyrol} & \multicolumn{2}{c}{Vienna} & \multicolumn{2}{c}{Overall}\\
\cline{2-13}
{} & IoU & Acc & IoU & Acc & IoU & Acc & IoU & Acc & IoU & Acc & IoU & Acc\\
\hline
FCN-8s & 50.28& 92.30& 53.89& 87.24& 32.09& 98.52& 56.40& 95.84& 62.75& 88.30& 56.19& 92.44\\
U-Net & \textbf{78.62}& \textbf{96.89}& \textbf{70.39}& \textbf{92.89}& \textbf{66.26}& \textbf{99.27}& 70.93& 97.71& 78.28& 93.85& 74.79& \textbf{96.12}\\
SegNet & 70.60& 94.74& 64.81& 89.72& 60.55& 98.89& 71.41& 97.28& 74.97& 91.79& 70.10& 94.48\\
DeepLab & 76.65& 96.54& 69.39& 92.56& 65.78& 99.24& \textbf{75.01}& \textbf{97.98}& \textbf{79.24}& \textbf{94.06}& \textbf{74.86}& 96.08\\
PSPNet & 71.69& 95.73& 66.67& 91.62& 63.08& 99.18& 72.07& 97.72& 76.49& 93.12& 71.67& 95.47\\
\hline
\end{tabular}
\end{center}
\label{table2}
\end{table}

\textbf{Segmentation Performance}    To evaluate the segmentation performance of methods, we use the accuracy and the Intersection over Union (IoU) which is more suitable for evaluating the unbalanced dataset. We evaluate the methods on the overall test set and for each city separately. Table~\ref{table2} shows that U-Net and DeepLab perform better than other models. More specifically, U-Net achieves the best result in the former three cities, and DeepLab is better in the latter two cities. For the overall test set, the IoU of DeepLab is better. It shows that DeepLab may be suitable for the unbalanced datasets which is more common in the real world environment.

\begin{figure}[h]
\begin{center}
   \includegraphics[width=0.8\linewidth]{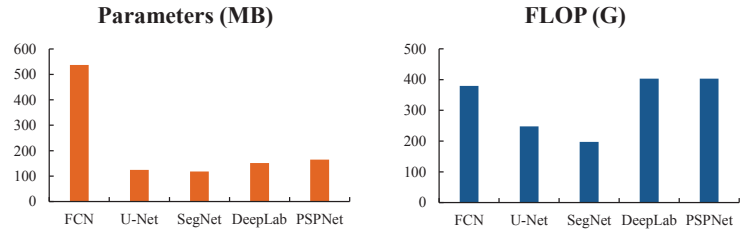}
\end{center}
   \caption{The parameters (left) are calculated on VGG based models. The FLOP (right) is measured on $512 \times 512$ inputs, except that DeepLab is $513 \times 513$ because of the atrous convolution.}
\label{bar_result}
\end{figure}

\textbf{Parameters}    Figure~\ref{bar_result} (left) shows parameters of each model. We find that FCN is the largest model and SegNet is the smallest one. Note that this indicator is closely related to storage resource and transmission capability of satellites.

\textbf{Resource Consumption}    Figure~\ref{bar_result} (right) illustrates floating point operations (FLOP) which can display the resource consumption of the model. SegNet and U-Net have smaller FLOP for Encoder-Decoder structure. And other models have larger FLOP for Multi-Scale representation structure.

\begin{figure}[h]
\begin{center}
   \includegraphics[width=0.8\linewidth]{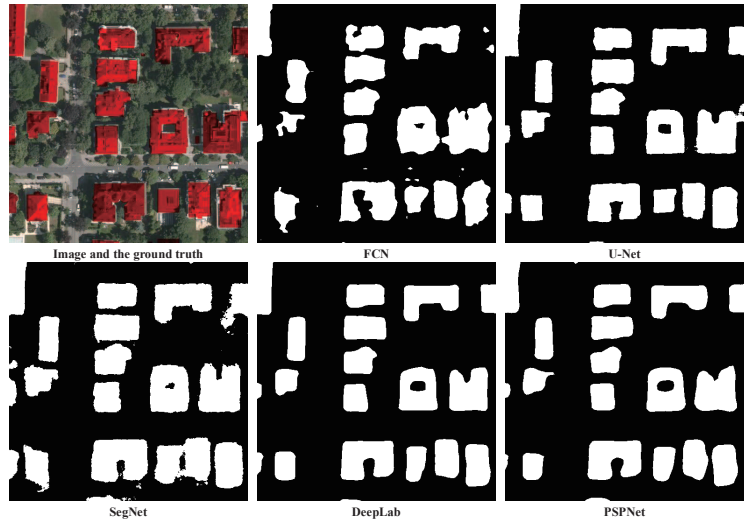}
\end{center}
   \caption{Segmentation results of Inria aerial image labeling dataset by using five methods.}
\label{seg_result}
\end{figure}

\textbf{Edge Prediction}    Figure~\ref{seg_result} shows several image segmentation results of the five methods. We observe that U-net and DeepLab do well in predicting edge which helps to restore the shape of targets. The edge predicted by SegNet is also sharp, but there are many points which are misclassified.

\section{On-orbit semantic segmentation strategy for TianZhi-2}
Due to the complexity of remote sensing image processing and limited energy on satellites, it is very necessary for us to consider segmentation performance and resource consumption simultaneously. Based on experimental results in Section 3, we propose that DeepLab is more suitable for on-orbit semantic segmentation. We propose a strategy to segment remote sensing images on the satellite as follows:

\textbf{Improve the semantic segmentation performance for remote sensing images.}
First, because the target distribution in remote sensing images is uneven and irregular, the object region extraction before segmentation will guide it and improve its performance.
Second, due to there are many irregular-shaped targets in remote sensing images, optimizing the target edge prediction can further improve the semantic segmentation performance.

\textbf{Reduce resource consumption by model pruning.}
In order to reduce the resource occupation on the satellite, we can use methods like $l_1$-norm to prune the model while keeping the performance levels almost intact and even better.

By using methods above, on-orbit semantic segmentation for remote sensing images can be achieved. After satellite capturing images, our image cropping method splits the images into smaller patches. The model takes these patches as input and outputs corresponding semantic segmentation results. Based on these results, there are many useful production such as the object image tile, target location and language description. They will greatly reduce the data transmission cost of satellite, which are powerful solutions for the limitation of satellite bandwidth.


\section{Conclusions}
In this paper, we make a research of deep learning based semantic segmentation methods on remote sensing images. We conduct a detailed comparative analysis of their structures and evaluate their performances using the same public dataset.
The experimental result shows that DeepLab does well in accuracy, parameters and resource consumption. By using this model, we propose a strategy for on-orbit semantic segmentation in TianZhi-2. We will further improve the method for the satellite environment, and strive to apply it to Tianzhi-2 which will be launched soon.

%
%

\end{document}